%% file: iclr2026_conference.tex
\documentclass{article} 
\usepackage{iclr2026_conference,times}

\input{math_commands.tex}

\usepackage{hyperref}
\usepackage{url}
\usepackage{graphicx}
\usepackage{booktabs}
\usepackage{multirow}
\usepackage{adjustbox}
\usepackage{subcaption}

\title{VDRive: Leveraging Reinforced VLA and Diffusion Policy for End-to-end Autonomous Driving}

\author{Ziang Guo \thanks{\emph{This work is done during Ziang Guo's internship in Suzhou Automotive Research Institute of Tsinghua University}} \\
Suzhou Automotive Research Institute \\of Tsinghua University \\
\texttt{slxx5237@gmail.com} \\
\And
Zufeng Zhang \\
Suzhou Automobile Research Institute \\ of Tsinghua University \\
\texttt{zhangzufeng@tsari.tsinghua.edu.cn} \\
}

\iclrfinalcopy 
\begin{document}

\maketitle

\begin{abstract}
In autonomous driving, dynamic environment and corner cases pose significant challenges to the robustness of ego vehicle's state understanding and decision making. We introduce VDRive, a novel pipeline for end-to-end autonomous driving that explicitly models state-action mapping to address these challenges, enabling interpretable and robust decision making. By leveraging the advancement of the state understanding of the Vision Language Action Model (VLA) with generative diffusion policy-based action head, our VDRive guides the driving contextually and geometrically. Contextually, VLA predicts future observations through token generation pre-training, where the observations are represented as discrete codes by a Conditional Vector Quantized Variational Autoencoder (CVQ-VAE). Geometrically, we perform reinforcement learning fine-tuning of the VLA to predict future trajectories and actions based on current driving conditions. VLA supplies the current state tokens and predicted state tokens for the action policy head to generate hierarchical actions and trajectories. During policy training, a learned critic evaluates the actions generated by the policy and provides gradient-based feedback, forming an actor-critic framework that enables a reinforcement-based policy learning pipeline. Experiments show that our VDRive achieves state-of-the-art performance in the Bench2Drive closed-loop benchmark and nuScenes open-loop planning.
\end{abstract}

\begin{figure*}[!h]
    \centering
    \begin{subfigure}{.47\linewidth}
        \centering
        \includegraphics[width=0.86\linewidth]{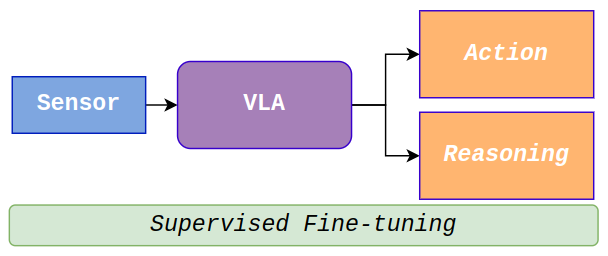}
        \caption{E2E supervised fine-tuning of VLA.}
    \end{subfigure}
    \begin{subfigure}{.47\linewidth}
        \centering
        \includegraphics[width=0.86\linewidth]{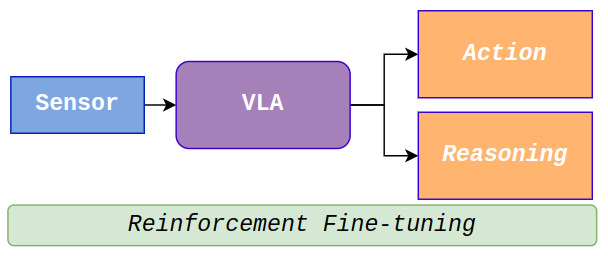}
        \caption{E2E reinforcement fine-tuning of VLA.}
    \end{subfigure}
    \begin{subfigure}{.47\linewidth}
        \centering
        \includegraphics[width=0.86\linewidth]{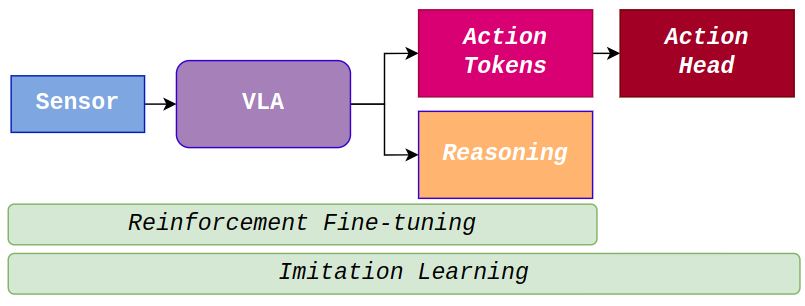}
        \caption{Reinforcement fine-tuning of VLA and a \\ unified IL-based E2E training.}
    \end{subfigure}
    \begin{subfigure}{.47\linewidth}
        \centering
        \includegraphics[width=0.86\linewidth]{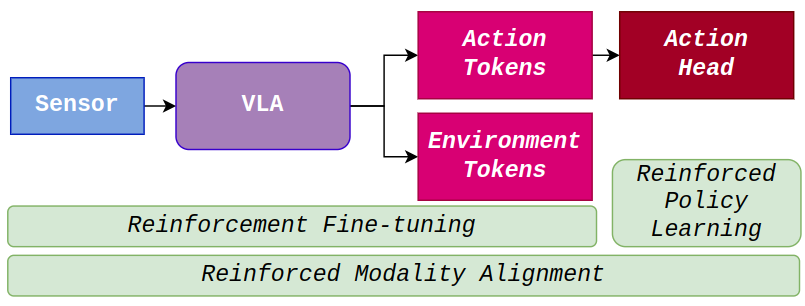}
        \caption{\textbf{Ours.} Reinforcement fine-tuning of VLA \\ and reinforced policy learning on head.}
    \end{subfigure}
    
    \caption{Comparison of VLA-based autonomous driving.}
    \label{fig:dev}
\end{figure*}

\section{Introduction}
\input{Intro}

\section{Related Work}
\input{Related}

\section{Dataset Construction}
\input{Data}

\section{Proposed Methodology}
\input{Method}

\section{Experiments}
\input{Exp}

\section{Conclusion}

In this work, we propose VDRive, a reinforced framework that aims to align sensor data space and action space through VLA's tokenization and diffusion policy's conditioning, addressing the challenges arising from dynamic driving environment and the infeasible action output. We combined real-world and synthetic datasets and performed reinforcement learning fine-tuning to the VLA and reinforced policy training to the diffusion model. In both open-loop and closed-loop evaluations, VDRive achieves state-of-the-art performance.

\textbf{Limitations.} To achieve the best performance, our framework involves multiple rounds of training and evaluation on VLA and policy head respectively, leading to extensive development and optimization. 

\textbf{Future work.} Thus, in the future, an E2E pipeline must be developed to enable stream gradient optimization and unified policy learning.

\bibliography{iclr2026_conference}
\bibliographystyle{iclr2026_conference}


\end{document}

%% file: math_commands.tex
\usepackage{amsmath,amsfonts,bm}

\def\eqref#1{equation~\ref{#1}}

\def\1{\bm{1}}

\DeclareMathAlphabet{\mathsfit}{\encodingdefault}{\sfdefault}{m}{sl}
\SetMathAlphabet{\mathsfit}{bold}{\encodingdefault}{\sfdefault}{bx}{n}

\newcommand{\E}{\mathbb{E}}

\newcommand{\R}{\mathbb{R}}

\DeclareMathOperator*{\argmin}{arg\,min}

%% file: Intro.tex
Over time, the end-to-end (E2E) autonomous driving paradigm has made significant progress \citep{xu2025knowledge, guidance, jiang2025flowdrive}. In the data-driven era, imitation learning (IL) has emerged as the dominant strategy: trajectories collected from expert drivers are used to supervise policy learning through behavioral cloning or to shape reward signals through inverse reinforcement learning and offline reinforcement learning \citep{liu2025reinforced, gao2025rad}. As the development of Large Language Models (LLMs) accelerates, the knowledge-driven paradigm has also advanced markedly \citep{guo2025vdt, xu2024vlmad}. LLMs are able to receive multimodal scene descriptions, convert them into semantically rich prompts, and generate high-level plans or low-level control. Critically, the knowledge encoded in LLM provides zero-shot generalization to rare traffic events and enables continual adaptation through contextual prompting, without costly recollection of on-road expert demonstrations \citep{jiang2025VLAsurvey}.

\par However, the E2E paradigm relies on multimodal perception, fusion, and alignment \citep{hwang2024emma}, where it aims to map raw sensor data directly to control actions through a unified learnable pipeline. A fundamental challenge lies in the significant representation gap between the high-dimensional, heterogeneous sensor data space (e.g., images, point clouds, audio) and the low-dimensional, structured action space (e.g., steering, throttle, braking). This gap is further exacerbated by semantic and temporal misalignment across modalities, as the features of the raw data operate at different levels of granularity \citep{li2024multimodal}.

\par Starting from this observation, a key motivation emerges: to bridge this gap through consistent, aligned multimodal representations that preserve both geometric and contextual features. Rather than treating modality fusion as a downstream concatenation or late stage merging, we advocate for a reinforced alignment strategy that harmonizes geometric and contextual features early in the pipeline, ensuring a coherent and interpretable data flow from sensing to action. 

\par In knowledge-driven approaches, Vision-Language-Action (VLA) models are effective in multimodal alignment, enabling action generation conditioned on visual observations and natural language prompts \citep{din2025VLAmani}. However, in autonomous driving, VLA needs to complete ambiguous or high-level goals, but the action output of VLA may lack the temporal coherence and fine-grained control required for continuous real-time driving \citep{xie2025vlmready}. To this end, we use a reinforced diffusion policy head that refines the coarse action proposals of the VLA into optimized and executable actions. The diffusion policy is jointly trained on the dataset used to fine-tune the VLA and the predictions by the fine-tuned VLA itself, enabling a consistent and synergistic learning framework.

\par With these insights, we propose VDRive, an E2E paradigm that bridges states and actions through reinforced VLA and diffusion policy with a tokenized state-action representation in which high-dimensional sensor inputs are quantized into discrete observation tokens, aligned with action tokens in a unified latent space. In VDRive, front-view images are provided to the VLA, which predicts the discrete future observation tokens along with the future trajectory, actions, and navigation commands. The diffusion policy head generates hierarchically denoised actions based on the states from the asynchronous input of current driving conditions and VLA's predictions. Finally, a joint refinement head optimizes the trajectory output by the dynamic guidance of asynchronous action predictions. Compared to other paradigms in Fig. \ref{fig:dev}, VDRive presents a fully offline reinforcement learning pipeline on both VLA and the diffusion policy head, aiming at the efficient alignment of the sensor and action space. Our contributions in this paper are summarized as follows: 
\begin{itemize}
    \item We introduce a novel pipeline, VDRive, which employs VLA and diffusion policy to geometrically and contextually instruct the driving, generating hierarchical actions and trajectories with reinforced modality alignment. For VLA, we first perform visual token pre-training for VLA's future observation token prediction. Then we reinforcedly fine-tune the VLA with a preference dataset of vision-action pairs. For the diffusion policy head, rule-based and learned rewards reinforce hierarchical action generation, while the refinement head optimizes the trajectory output guided by these hierarchical actions.
    \item We construct a preference dataset for VLA reinforcement learning fine-tuning, where the chosen and rejected vision-action pairs are based on the nuScenes and the processed Bench2Drive dataset. For reinforcement training of the diffusion policy head, we construct an offline reward dataset that builds on the preference dataset and the predictions of the fine-tuned VLA. Rewards are provided through a combination of rule-based evaluation and behavior ratings from an expert Vision Language Model (VLM). The constructed dataset will be publicly available.
    \item In the nuScenes open-loop planning evaluation, VDRive achieved $0.29$ m on average L2 errors and $18\%$ on average collision rate. In the Bench2Drive benchmark, VDRive achieved a Driving Score of $66.25$ and a Success Rate of $50.51\%$.
\end{itemize}

%% file: Related.tex
\subsection{VLA}

For VLA, both explicit and implicit reasoning present promising results. However, explicit reasoning involves manual prompt design and poses a computational load on real-time deployment. Implicit reasoning may introduce difficulty in long-tail or challenging scenarios due to physically infeasible action outputs. The potential of VLA to serve as a foundation for end-to-end driving systems have been demonstrated by OpenDriveVLA, which unifies semantic reasoning of cross-modality and 3D instance-aware trajectory planning. The model leverages the pre-trained VLM to generate reliable driving actions conditioned on 3D environmental perception, ego states, and driver commands together \citep{zhou2025opendrivevla}. AutoVLA employs supervised fine-tuning with dual thinking modes (fast and slow) and reinforcement fine-tuning based on Group Relative Policy Optimization (GRPO) to improve planning performance, efficiency, and adaptive reasoning in various scenarios \citep{zhou2025autovla}. DiffVLA is a novel hybrid sparse-dense diffusion policy guided by a VLM for end-to-end autonomous driving. DiffVLA addresses challenges in existing methods by improving trajectory generation guidance through deep interaction among agent, map instances, and VLM output \citep{jiang2025diffvla}. Considering the efficiency of VLA's deployment, FastDriveVLA employs a plug-and-play visual token pruner called ReconPruner, trained via MAE-style pixel reconstruction with an adversarial foreground-background strategy to prioritize relevant foreground information \citep{cao2025fastdrivevla}. IRL-VLA presents a three-stage approach: pretraining a VLA policy with imitation learning, building a lightweight Reward World Model (RWM) through inverse reinforcement learning for efficient reward computation, and fine-tuning the VLA policy using Proximal Policy Optimization (PPO) with RWM guidance \cite{jiang2025irlvla}. This allows for scalable and effective close-loop reinforcement learning without relying on computationally expensive simulators. Drawing on the above, VLA's reasoning ability is the core of these paradigms. With tokenization of sensor space, our VDRive unites efficient implicit and explicit representations as contextual and geometric reasoning to guide the driving.

\subsection{Diffusion Policy}

Diffusion model-based approaches have proven their value in policy learning tasks \citep{chi2023diffusionpolicy, yang2024diff-es}. Since the advancement of diffusion models, they have gained recognition as a cornerstone in the field of generative modeling \citep{peebles2023DiT}. Conditioned diffusion models extend vanilla diffusion models by incorporating additional information during the generation process, while latent diffusion models improve computational efficiency and sample quality by operating in a compressed latent space \citep{rombach2022latentDiff}. Diffusion models have shown a strong potential to generate high-quality data in various modalities and to improve the representation of complex data structures \citep{yang2023diffrep}. In robotics, Cheng Chi et al. \citep{chi2023diffusionpolicy}, proposing a multimodal probabilistic action representation, considered the generation of robot action as a conditional diffusion denoising process. RDT-1B used a scalable Transformer backbone combined with diffusion models to capture the complexity and multimodality of bimanual actions, using diffusion models as a foundation model to effectively represent the multimodality inherent in bimanual manipulation tasks \citep{liu2024rdt-1b}. In autonomous driving, the objectives of control policies are typically ill-defined and dynamically evolving rather than statically specified. The DIVER framework integrates diffusion models and reinforcement learning to address mode collapse in imitation learning for end-to-end autonomous driving. It proposes a Policy-Aware Diffusion Generator (PADG) that generates various trajectory candidates by conditioning on map elements, surrounding agents, and multiple reference ground truth trajectories \citep{song2025breaking}. As a planner, diffusion models excel in generating a wide distribution of diverse trajectories, while reinforcement learning can then optimize the generated trajectories based on specific diversity and safety rewards, effectively guiding the diffusion process to produce physically plausible and robust driving behaviors. Our framework jointly leverages diffusion policy and VLA via training data interaction and reinforcement learning, aiming at the efficient alignment of the sensor and action space.

%% file: Data.tex
\subsection{Pre-training and RL Fine-tuning of VLA}

Instead of a full range of raw images without specific attention, we aim to enable VLA's visual understanding in driving tasks by segmenting the drivable areas and lane borders from the front view. At the same time, the designed rewards $R_h$ are also derived from the segmentation masks and the ego states. We introduce more details of the designed rule-based rewards $R_h$ in Section 3.2.

\par Based on the nuScenes dataset, we generate risky scenes $I'_r$ using Vista's trajectory control for each sample \citep{caesar2020nuscenes, gao2024vista}. Using CVQ-VAE with conditioning on the corresponding trajectories, we obtain the discrete codes of segmented raw scenes and generated scenes. Similarly to FSDrive \citep{zeng2025futuresightdrive}, discrete codes are added to the vocabulary of the VLA tokenizer. We then construct the token generation pre-training dataset with input of prompts and front-view images and output of discrete codes. In the CARLA simulator, we process the Bench2Drive-base dataset to obtain risky scenes along with segmentation masks and rewards \citep{jia2024bench2drive}.

\par
\begin{figure}[h]
    \centering
    \includegraphics[width=0.7\linewidth]{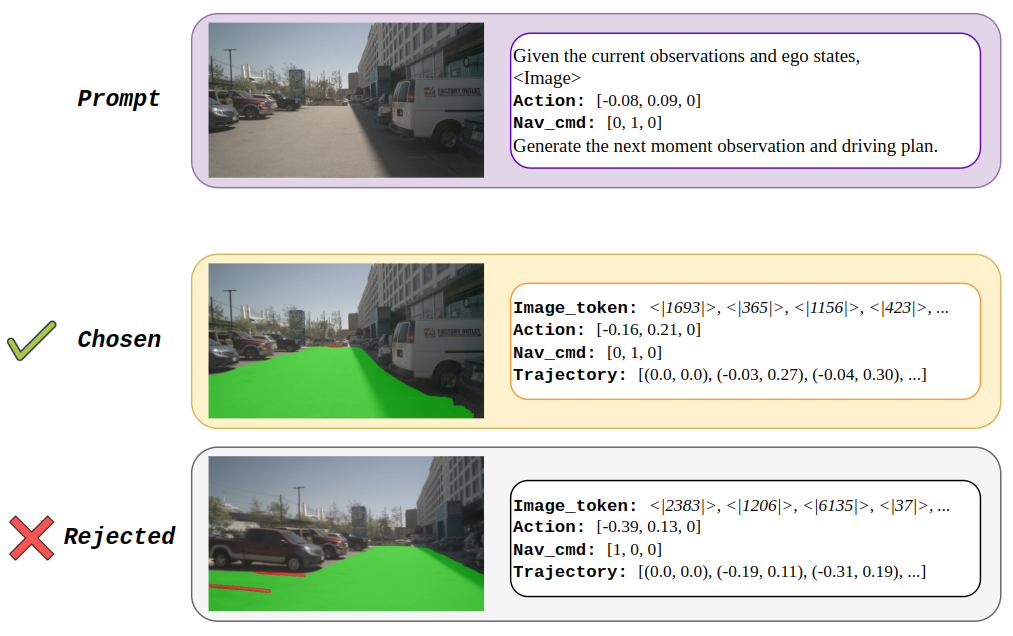}
    \caption{Example of our reinforcement learning fine-tuning dataset.}
    \label{fig:dataset_example}
\vspace{-5mm}
\end{figure}

 To fine-tune the VLA, we build a visual reinforcement learning dataset with the chosen and rejected output, where the chosen output is paired with safe scenes annotated by nuScenes and Bench2Drive and the rejected output is constructed with Vista-generated and selected Bench2Drive risky scenes. Fig. \ref{fig:dataset_example} shows an example of our created dataset.

\subsection{Diffusion Policy Head Training}

To extend the policy head training dataset with diverse reward signals, we combine annotated, Vista-generated, and VLA-predicted state-action pairs to compute the proposed rewards. In reward construction, we design a hybrid reward that combines the rating of an expert VLM Qwen2.5-VL-72B and the rules based on the drivable area and the ego states \citep{bai2025qwen2}. 

\textbf{Rule-based rewards.} Given binary drivable area segmentation mask $D \in \{0, 1\}^{H \times W}$ and trajectories in the camera coordinates $T = \{ (x_t, y_t) \}_{t=1}^N$, where $ H, W $ are the height and width of the image, respectively. The off-road indicator for each point on the trajectory is defined as

\begin{equation}
    \text{OffRoad}_t = 
\begin{cases}
0 & \text{if } D_{\lfloor y_t \rfloor, \lfloor x_t \rfloor} = 1 \\
1 & \text{otherwise}.
\end{cases}
\end{equation}

Total off-road penalty over the trajectory:
\begin{equation}
    P_{\text{off}} = \sum_{t=1}^N \text{OffRoad}_t.
\end{equation}

In addition, for each row of drivable masks, let $S_y = \{x \in [0,W] \mid D_{y,x} = 1\}$ be a set of drivable columns in the row $y \in [0,H]$. $\mu_y = \frac{1}{|S_y|} \sum_{x \in S_y} x$ is the mean x-coordinate of the drivable masks in row $y$. Then, for a trajectory $(x_t,y_t)$, define its normalized lateral deviation from the center:

\begin{equation}
    d_t = \frac{|x_t - \mu_{\lfloor y_t \rfloor}|}{(max(S_{\lfloor y_t \rfloor}) - min(S_{\lfloor y_t \rfloor})) / 2}.
\end{equation}

The centering award is computed as

\begin{equation}
    R_{\text{center}} = \frac{1}{N} \sum_{t=1}^N \exp\left(-\alpha \cdot d_t^2\right),
\end{equation}

where $\alpha > 0$ is a scaling factor.

The rule-based reward is finally combined as

\begin{equation}
    R_h = R_{\text{center}} \cdot \mathbb{I}(P_{\text{off}} = 0) - \beta \cdot \mathbb{I}(P_{\text{off}} > 0),
\end{equation}

where $\beta$ is a large negative value and $\mathbb{I}$ is the indicator function.

\begin{figure}[h]
    \centering
    \includegraphics[width=0.9\linewidth]{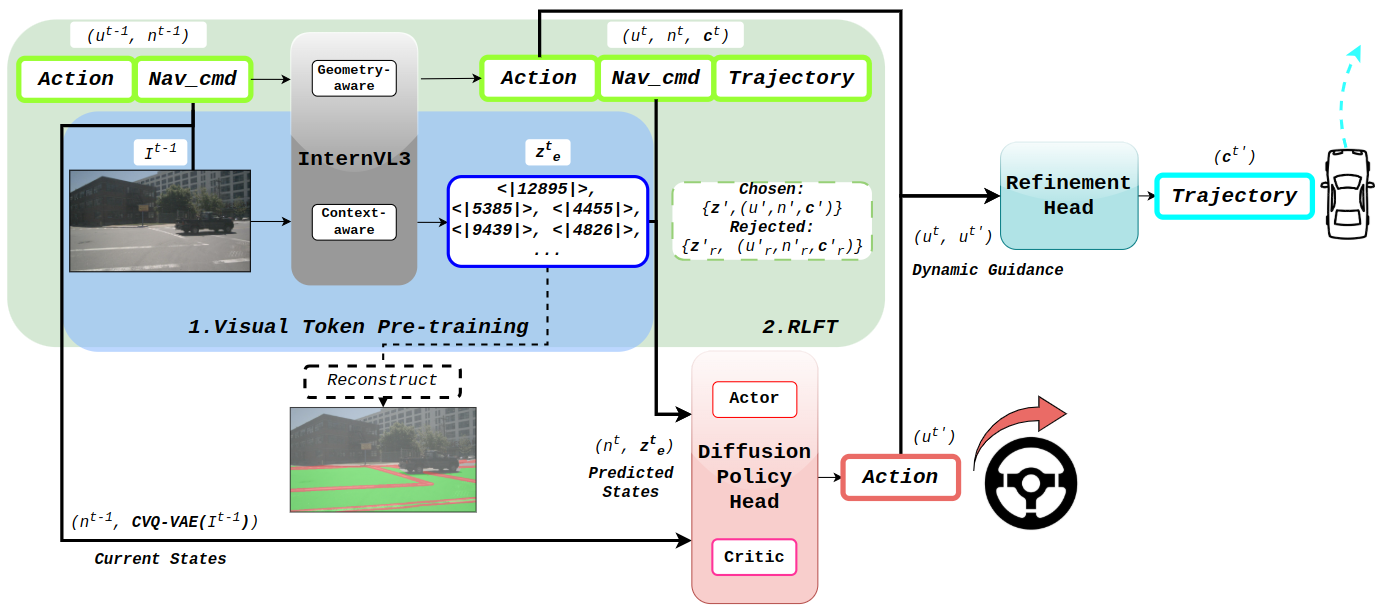}
    \caption{\textbf{Proposed VDRive framework.} A reinforced framework empowering VLA’s prediction of segmented drivable future and policy heads' generation of hierarchical actions and trajectories. With reinforcement learning and discrete representation of sensor data, VDRive is trained to align context and geometry tokens in a low dimensional space.}
    \label{fig:overview}
\vspace{2mm}
\end{figure}

\textbf{Expert VLM-based rewards.} Using Qwen2.5-VL-72B as a traffic risk analyst, we feed it all the state-action pairs in our dataset to obtain the risk rating by demonstrating the classifications of driving safety. Given the distances to the detected objects and ego states in the scenes, the VLM is first demonstrated with safe scenes with a rating of 0 and risky scenes with a rating of 5, where the safe scenes have the longest distances to detected objects and continuous motions, while the risky scenes are identified with the shortest distances to detected objects and discontinuous motions. All VLM-based ratings are within the range $[0, 5]$, where the rewards are extracted as $R_a$.

Hence, the final reward is defined as

\begin{equation}
    R = \underbrace{\omega_h \cdot {R_h}}_{\text{Rule-based reward}} +\underbrace{\omega_a \cdot R_{a}}_{\text{VLM-based reward}},
\end{equation}

where $\omega_h, \omega_a$ are the weights for the final reward.

%% file: Method.tex
\subsection{Preliminary}

\textbf{CVQ-VAE.} The premise of the proposed reinforced modality alignment is the discretization of a continuous sensor data space \citep{mentzer2023finite}, where we train CVQ-VAE in the dataset that we combined as above. Given the conditioning variable $\mathbf{c} = \{ (x_t, y_t) \}_{t=1}^T \in \R^{T \times 2}$, the raw image input $\mathbf{I}$, and the segmented target $\mathbf{I_{\text{seg}}}$, CVQ-VAE first encodes the raw image and the trajectory into a continuous latent space:

\begin{equation}
    \mathbf{z}_e = \text{Enc}_\theta(\mathbf{I}, \mathbf{c}).
\end{equation}

Then each vector in $\mathbf{z}_e$ is mapped to the nearest code in a learnable codebook $\mathcal{C} = \{ e_1, ..., e_K\}$, $e_K \in \R^D$. For each spatial location:

\begin{equation}
    \mathbf{z}_q[i,j] = \mathbf{e}_k, \quad k = \arg\min_{k'} \| \mathbf{z}_e[i,j] - \mathbf{e}_{k'} \|^2.
\end{equation}

The total loss is defined as follows.

\begin{equation}
    \mathcal{L} = 
\underbrace{\mathcal{L}_{\text{recon}}(\mathbf{I}_{\text{seg}}, \hat{\mathbf{I}}_{\text{seg}})}_{\text{Reconstruction}} + 
\underbrace{\| \mathbf{z}_q - \text{sg}(\mathbf{z}_e) \|^2}_{\text{Codebook Loss}} + 
\beta \underbrace{\| \text{sg}(\mathbf{z}_e) - \mathbf{z}_q \|^2}_{\text{Commitment Loss}},
\end{equation}

where the reconstruction loss is represented as 

\begin{equation}
    \mathcal{L}_{\text{recon}} = -\sum_{i,j} \left[ \mathbf{I}_{\text{seg}}[i,j] \log \hat{\mathbf{I}}_{\text{seg}}[i,j] + (1 - \mathbf{I}_{\text{seg}}[i,j]) \log (1 - \hat{\mathbf{I}}_{\text{seg}}[i,j]) \right].
\end{equation}

$\text{sg}( \cdot )$ is the stop-gradient operator and $\beta > 0$ is the commitment cost.

\textbf{Diffusion policy learning.} Gathering the computed rewards and discrete codes from our annotated, Vista-generated, and VLA-predicted data, we formulate the offline dataset created for policy head training as $\mathcal{D} = \{ ( s, a, R, s' ) \}$. Following \citep{chen2024diffusionoffline}, we train policy head $\pi_\theta(a|s)$ to maximize the cumulative reward $\E{\sum_{t=0}^{\infty}\gamma^t R(s_t, a_t)}$ with the discount factor $\gamma^t \in [0, 1]$. The critic network for the policy head $\pi_\theta$ is defined as $Q^\pi(s_t, a_t) = \E_{a_{t+1}, a_{t+2}, ... \sim \pi}{\sum_{t=0}^{\infty}\gamma^t R(s_t, a_t)}$. As an actor network $\mu_\phi$, the diffusion model is trained to denoise the noisy input conditioning on the state-action pairs,

\begin{equation}
      \mathcal{L}(\phi)=\E_{t,\bm{\varepsilon}\sim \mathcal{N}(0,\bm{I}),(\bm{a}_0,\bm{s})\sim\mathcal{D}}[\|\mu_{\phi}(\bm{a}_t,t|\bm{s})-\bm{a}_0\|_2^2]
\end{equation}

where $a_t = \alpha_t a_0 + \sigma_t \epsilon$ is the noise addition process controlled by $\alpha$, $\sigma$ at each diffusion timestep, and $\epsilon$ is sampled from random Gaussian noise.

Combining with the critic network $Q^\pi$, the final objective of policy learning is 
\begin{equation}
    \pi = \argmin_{\pi_\theta} \mathcal{L_{\pi}}(\theta) = \mathcal{L}(\theta) + \mathcal{L}_q(\theta) = \mathcal{L}(\theta) - \omega_Q \cdot \E_{s \sim \mathcal{D}, a^0 \sim \pi_\theta} {Q^{\pi}(s, a^0)},
\end{equation}

where $\omega_Q$ is the normalization scale of the critic.

\subsection{Reinforced Modality Alignment}

In this section, we elucidate the proposed reinforced modality alignment as shown in Fig. \ref{fig:overview}. In the VDRive training pipeline, discrete codes $z_{seg} = \text{CVQ-VAE}(I_{seg}, \mathbf{c})$ and $z'_{seg} = \text{CVQ-VAE}(I'_{seg}, \mathbf{c}')$ conditioning on the trajectory $\mathbf{c} = \{ (x_t, y_t) \}_{t=1}^T \in \R^{T \times 2}$ and $\mathbf{c}' = \{ (x'_t, y'_t) \}_{t=1}^T \in \R^{T \times 2}$ of the segmented raw and Vista-generated images are first added to the InternVL3-8B-based VLA tokenizer vocabulary through visual token pretraining \citep{zhu2025internvl3}. During reinforcement learning fine-tuning of VLA, given the chosen output $(z^t_{seg}, [u^t_s, u^t_t, u^t_b], [n^t_1, n^t_2, n^t_3], \mathbf{c}^t)$ and the rejected output $(z'^t_{seg}, [u'^t_s, u'^t_t, u'^t_b], [n'^t_1, n'^t_2, n'^t_3], \mathbf{c}'^t)$, we expect that VLA generates

\begin{equation}
    (z^t_{seg}, [u^t_s, u^t_t, u^t_b], [n^t_1, n^t_2, n^t_3], \mathbf{c}^t) = VLA(I^{t-1}, [u^{t-1}_s, u^{t-1}_t, u^{t-1}_b], [n^{t-1}_1, n^{t-1}_2, n^{t-1}_3]),
\end{equation}

where $n = [n_1, n_2, \dots, n_k], n_k \in \{0, 1\}$ are the one-hot navigation commands and $u = [\text{steering}, \text{throttle}, \text{brake}] \in [-1,1] \times [0,1] \times [0,1]$ are the action signals. 

In diffusion policy training, as described in Section 4.1, the training process starts as follows. 

\begin{equation}
    ([u_s, u_t, u_b]) = \pi(z_{seg}, [u^0_s, u^0_t, u^0_b], [n_1, n_2, n_3]).
\end{equation}

During inference, the diffusion policy head generates hierarchically denoised actions by having the state input to VLA at time $t-1$ and VLA's state prediction at time $t$.


\subsection{Dynamics-guided Refinement}

With the dynamic guidance formed by the asynchronous action predictions of the VLA and the diffusion policy head, we optimize the trajectory output as follows.

\begin{equation}
    \hat{\mathbf{c}} = \mathrm{Decode} \left( \mathrm{TransformerEncoder} \left( \mathrm{ReLU} \left( \left[ \mathrm{Proj}(\mathbf{c}) + \mathbf{E}_{\mathrm{pos}},\; \mathrm{Tile}\left( \mathrm{MLP}(\mathbf{U}) \right) \right] \right) \right) \right) + \mathbf{c},
\end{equation}

where $\mathbf{U} = [\mathbf{u}_{t-1}, \mathbf{u}_t] \in \mathbb{R}^{2 \times 3}$ is the input of asynchronous actions and $\mathbf{E}_{\mathrm{pos}}$ is the learnable positional embeddings. During training, the refinement head is trained with action-trajectory pairs from our created dataset to enable trajectory optimization with learned dynamics.

%% file: Exp.tex

\subsection{Open-loop Experiments}

For open-loop evaluation, we used nuScenes, consisting of 1000 scenes with measurements from 1 spinning LiDAR, 6 cameras, 5 long-range radars, etc. The well-synchronized samples from the cameras, LiDAR, and radars are at 2 Hz. We compared our VDRive with other methods using the L2 error in meters and the collision rate as a percentage in Table \ref{tab:L2}. The average L2 error is computed as the distance between each waypoint in the planned trajectory and the corresponding waypoint in the ground truth trajectory. The collision rate is assessed by positioning an ego-vehicle bounding box at each waypoint along the planned trajectory and subsequently checking for any intersections with the ground truth bounding boxes of other objects.

\begin{table*}[h]
\caption{The open-loop planning results of our VDRive on nuScenes validation set.}
\centering
\begin{adjustbox}{width=0.88\linewidth}{
    \begin{tabular}{l|l|c|c|c|c|c|c|c|ccc}
        \toprule
        \multirow{2}{*}{No.} & \multirow{2}{*}{Methods} & \multicolumn{4}{c|}{L2 (m) $\downarrow$} & \multicolumn{4}{c}{Collision Rate (\%) $\downarrow$} \\
         & & 1s & 2s & 3s & Avg. & 1s & 2s & 3s & Avg. \\
        \midrule
        1 & FF~\citep{hu2021safeff} & 0.55 & 1.20 & 2.54 & 1.43 & 0.06 & 0.17 & 1.07 & 0.43 \\
        2 & EO~\citep{kh2022differentiable} & 0.67 & 1.36 & 2.78 & 1.60 & 0.04 & 0.09 & 0.88 & 0.33 \\
        3 & ST-P3~\citep{hu2022st-p3} & 1.33 & 2.11 & 2.90 & 2.11 & 0.23 & 0.62 & 1.27 & 0.71 \\
        4 & UniAD~\citep{hu2023uniAD} & 0.48 & 0.96 & 1.65 & 1.03 & 0.05 & 0.17 & 0.71 & 0.31 \\
        5 & GPT-Driver~\citep{mao2023gpt} & 0.27 & 0.74 & 1.52 & 0.84 & 0.07 & 0.15 & 1.10 & 0.44 \\
        6 & VLP-UniAD~\citep{pan2024vlp} & 0.36 & 0.68 & 1.19 & 0.74 & 0.03 & 0.12 & 0.32 & 0.16 \\
        7 & RDA-Driver~\citep{huang2024rda} & 0.23 & 0.73 & 1.54 & 0.80 & \textbf{0.00} & 0.13 & 0.83 & 0.32 \\
        8 & DriveVLM~\citep{tian2024drivevlm} & 0.18 & 0.34 & 0.68 & 0.40 & 0.10 & 0.22 & 0.45 & 0.27 \\
        9 & HE-Drive-B~\citep{wang2024hedrive} & 0.30 & 0.56 & 0.89 & 0.58 & \textbf{0.00} & \textbf{0.03} & \textbf{0.14} & \textbf{0.06} \\
        10 & ReAL-AD~\citep{lu2025real-ad} & 0.30 & 0.48 & 0.67 & 0.48 & 0.07 & 0.10 & 0.28 & 0.15 \\
        
        \midrule
        10 & \textbf{Ours} & \textbf{0.12} & \textbf{0.26} & \textbf{0.50} & \textbf{0.29} & 0.03 & 0.16 & 0.36 & 0.18 \\

        \midrule
       
    \end{tabular}
}
\end{adjustbox}
\label{tab:L2}
\end{table*}

\subsection{Closed-loop Experiments}

We evaluated VDRive using Bench2Drive, a closed-loop evaluation in the CARLA simulator. As shown in Table \ref{tab:b2d}, our VDRive achieved a Driving Score of $66.25$, Success Rate of $50.51\%$ with Efficiency of $110.23$ and Comfortness of $22.90$. In Table \ref{tab:ability}, we report the Multi-ability results with Merging$(22.35)$, Overtaking$(48.23)$, Emergency Brake$(50.96)$, Give Way$(40.00)$, and Traffic Sign$(66.72)$. 

\begin{table}[tb!]
\centering
\small
\caption{The closed-loop evaluation of Bench2Drive. Avg. L2 is averaged over the predictions in 2 seconds under 2Hz. * denotes expert feature distillation. \label{tab:b2d}}
\begin{adjustbox}{width=0.98\linewidth}{
\begin{tabular}{l|c|c|c|c|c}
\toprule
\multirow{2}{*}{\textbf{Method}} & \textbf{Open-loop Metric} & \multicolumn{4}{c}{\textbf{Closed-loop Metric}} \\ \cmidrule{2-6} 
    &      Avg. L2 $\downarrow$     &  Driving Score $\uparrow$  & Success Rate(\%) $\uparrow$ & Efficiency $\uparrow$ & Comfortness $\uparrow$\\ \midrule
AD-MLP~\citep{zhai2023ADMLP}            & 3.64              &  18.05     &  0.00  & 48.45 &   22.63  \\ 
UniAD-Tiny~\citep{hu2023planning}          &  0.80       &  40.73    & 13.18 & 123.92 & 47.04   \\
UniAD-Base~\citep{hu2023planning}             &  0.73        &  45.81     & 16.36 & 129.21 & 43.58  \\
VAD~\citep{jiang2023vad}           &    0.91     &  42.35     & 15.00 & \textbf{157.94} & 46.01 \\
TCP*~\citep{wu2022trajectoryguided}    & 1.70    & 40.70     & 15.00  & 54.26 & \textbf{47.80}  \\ 
ThinkTwice*~\citep{jia2023thinktwice}        & 0.95   &  62.44     & 31.23  & 69.33 & 16.22  \\
DriveAdapter*~\citep{jia2023driveadapter}        & 1.01       &  64.22     & 33.08 & 70.22 & 16.01  \\ 
DriveTransformer~\citep{jia2025drivetransformer} & 0.62 &  63.46 &  35.01 & 100.64 & 20.78  \\ 
ReAL-AD~\citep{lu2025real-ad} & 0.84 &  41.17 &  11.36 & - & -  \\ 
CogAD~\citep{wang2025cogad} & - &  48.30 & 24.00  & 142.00  & 40.37 \\
\textbf{Ours} & \textbf{0.55} &  \textbf{66.25} &  \textbf{50.51} & 110.23 & 22.90  \\ \bottomrule

\end{tabular}}
\end{adjustbox}
\end{table}
\begin{table}[]
\centering
\caption{Multi-ability results of E2E-AD methods in Bench2Drive.  * denotes expert feature distillation.}\label{tab:ability}
\begin{adjustbox}{width=\linewidth}{
\begin{tabular}{l|ccccc|c}
\toprule
\multirow{2}{*}{\textbf{Method}} & \multicolumn{5}{c}{\textbf{Multi-ability} (\%) $\uparrow$}                                                                                                                \\ \cmidrule{2-7} 
                                 & \multicolumn{1}{c}{Merging} & \multicolumn{1}{c}{Overtaking} & \multicolumn{1}{c}{Emergency Brake} & \multicolumn{1}{c}{Give Way} & Traffic Sign & \textbf{Mean} \\ \midrule
AD-MLP~\citep{zhai2023ADMLP}        & 0.00        & 0.00            & 0.00        & 0.00           &  0.00    & 0.00         \\
UniAD-Tiny~\citep{hu2023planning}   & 7.04        & 10.00           & 21.82       &  20.00         & 14.61    & 14.69        \\ 
UniAD-Base~\citep{hu2023planning}       & 12.16           & 20.00       &  23.64         & 10.00    & 13.89 & 15.94       \\ 
VAD~\citep{jiang2023vad}            & 7.14        & 20.00          & 16.36       &  20.00         & 20.22    & 16.75       \\
TCP*~\citep{wu2022trajectoryguided}        & 17.50        & 13.63           & 20.00        &  10.00         & 6.81     & 13.59        \\
ThinkTwice*~\citep{jia2023thinktwice}      & 13.72       &  22.93          & 52.99       &  \textbf{50.00}        & 47.78    & 37.48        \\ 
DriveAdapter*~\citep{jia2023driveadapter}      & 14.55       & 22.61          & \textbf{54.04}      &  \textbf{50.00}         & 50.45   & 38.33        \\   
DriveTransformer~\citep{jia2025drivetransformer} & 17.57 & 35.00 &  48.36 & 40.00 &  52.10 & 38.60 \\ 
\textbf{Ours} & \textbf{22.35} & \textbf{48.23} &  50.96 & 40.00 &  \textbf{66.72} & \textbf{45.65} \\ \bottomrule
\end{tabular}}
\end{adjustbox}
\end{table}

\subsection{Qualitative Results}

\begin{figure*}
    \centering
    \begin{subfigure}{.3\linewidth}
        \centering
        \includegraphics[width=\linewidth]{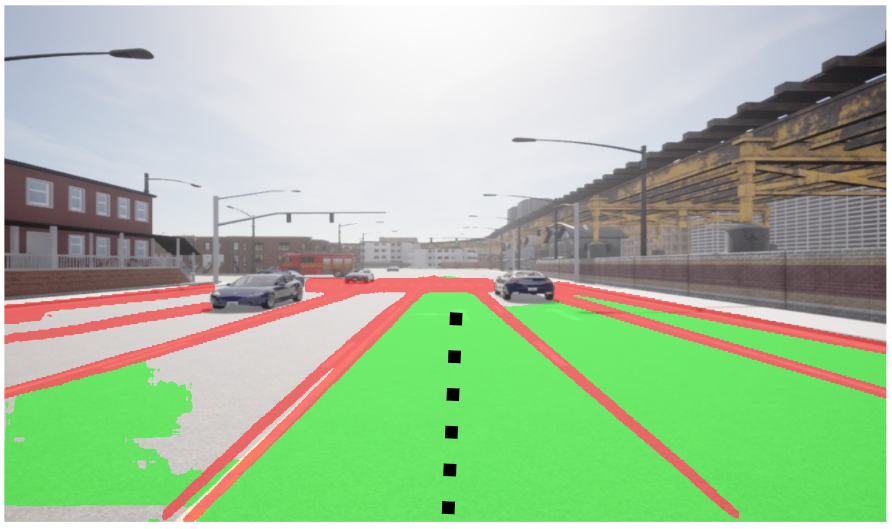}
    \end{subfigure}
    \begin{subfigure}{.3\linewidth}
        \centering
        \includegraphics[width=\linewidth]{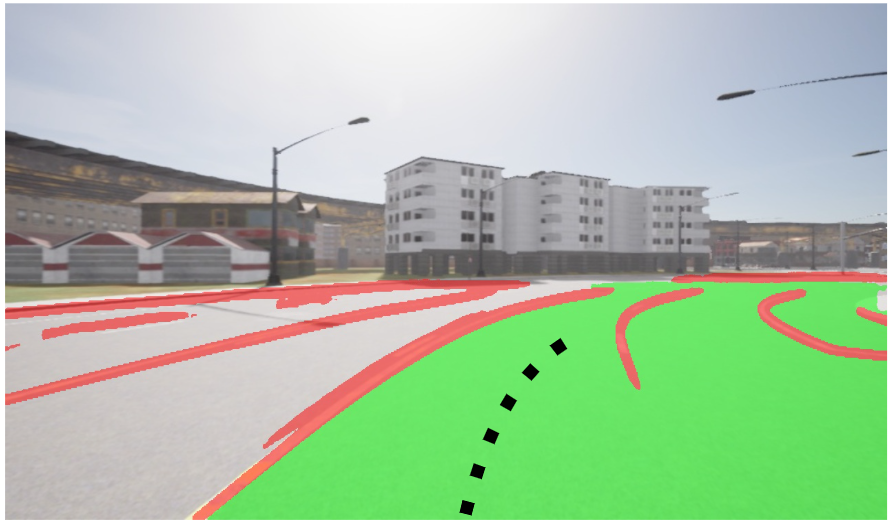}
    \end{subfigure}
    \begin{subfigure}{.3\linewidth}
        \centering
        \includegraphics[width=\linewidth]{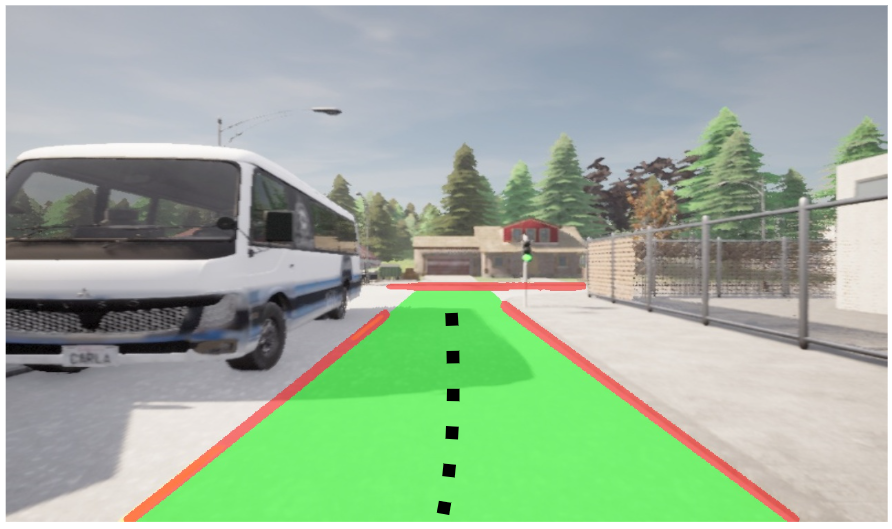}
    \end{subfigure}

    \begin{subfigure}{.3\linewidth}
        \centering
        \includegraphics[width=\linewidth]{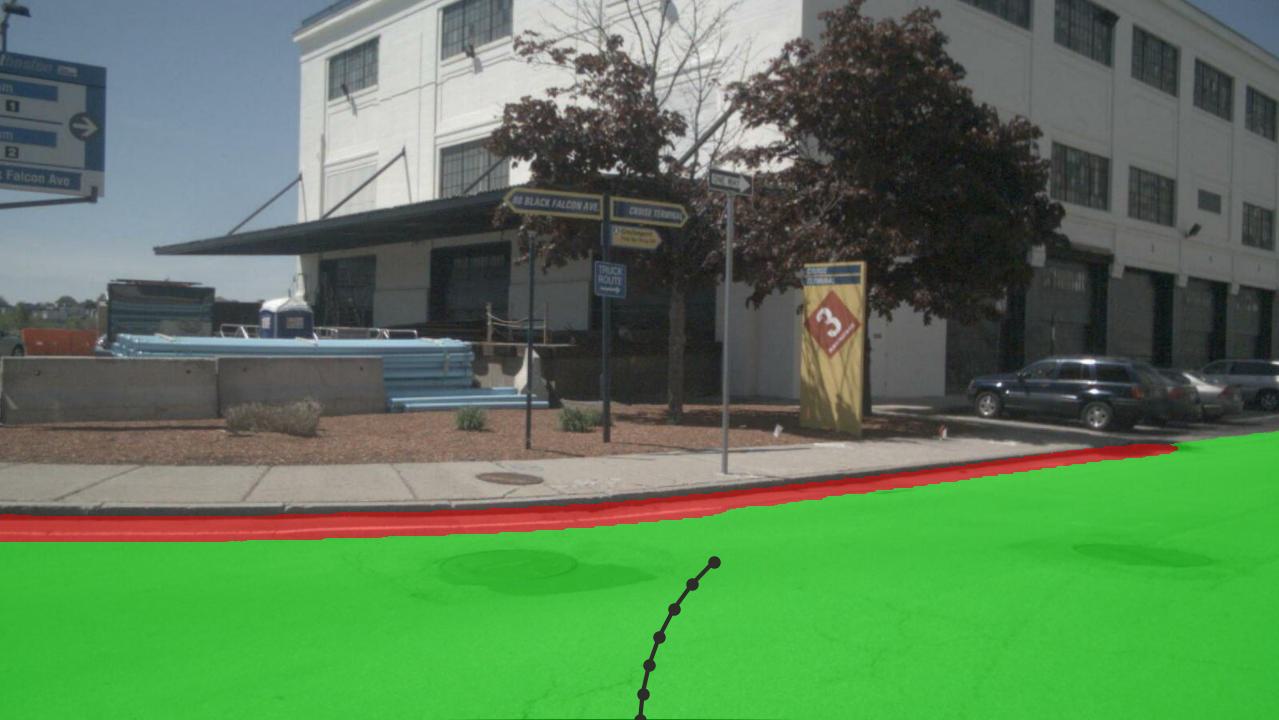}
    \end{subfigure}
    \begin{subfigure}{.3\linewidth}
        \centering
        \includegraphics[width=\linewidth]{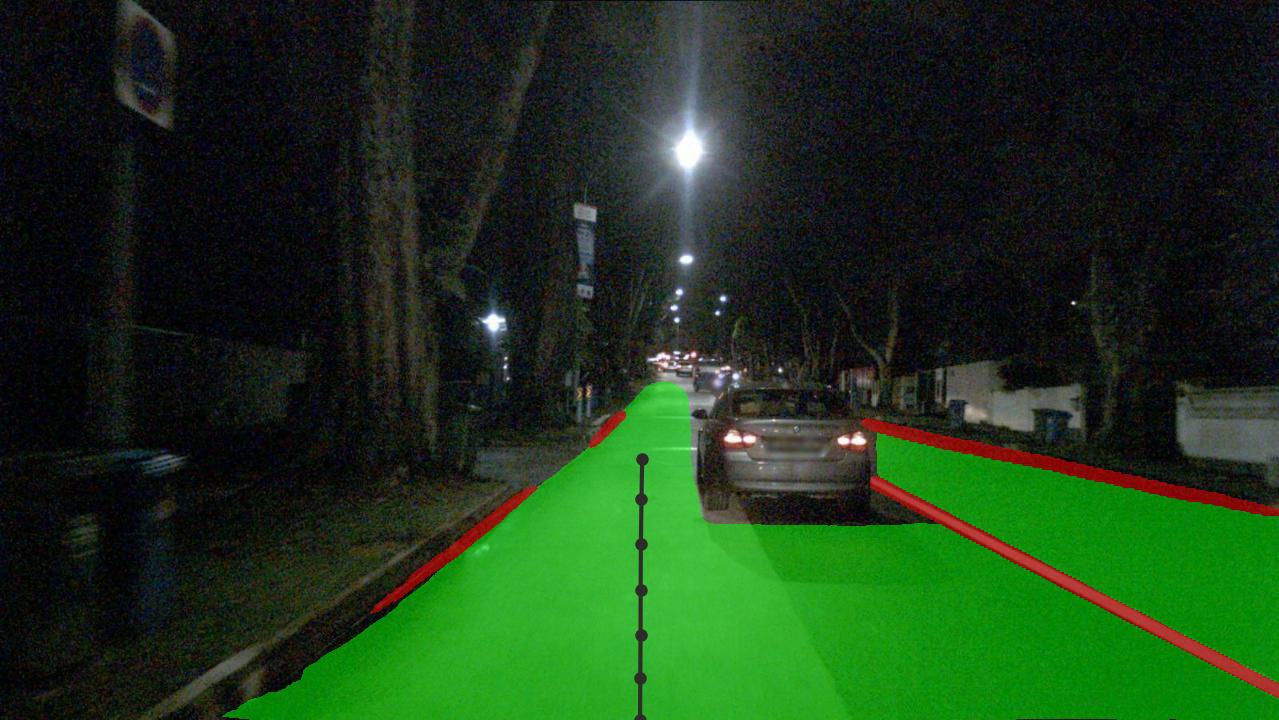}
    \end{subfigure}
    \begin{subfigure}{.3\linewidth}
        \centering
        \includegraphics[width=\linewidth]{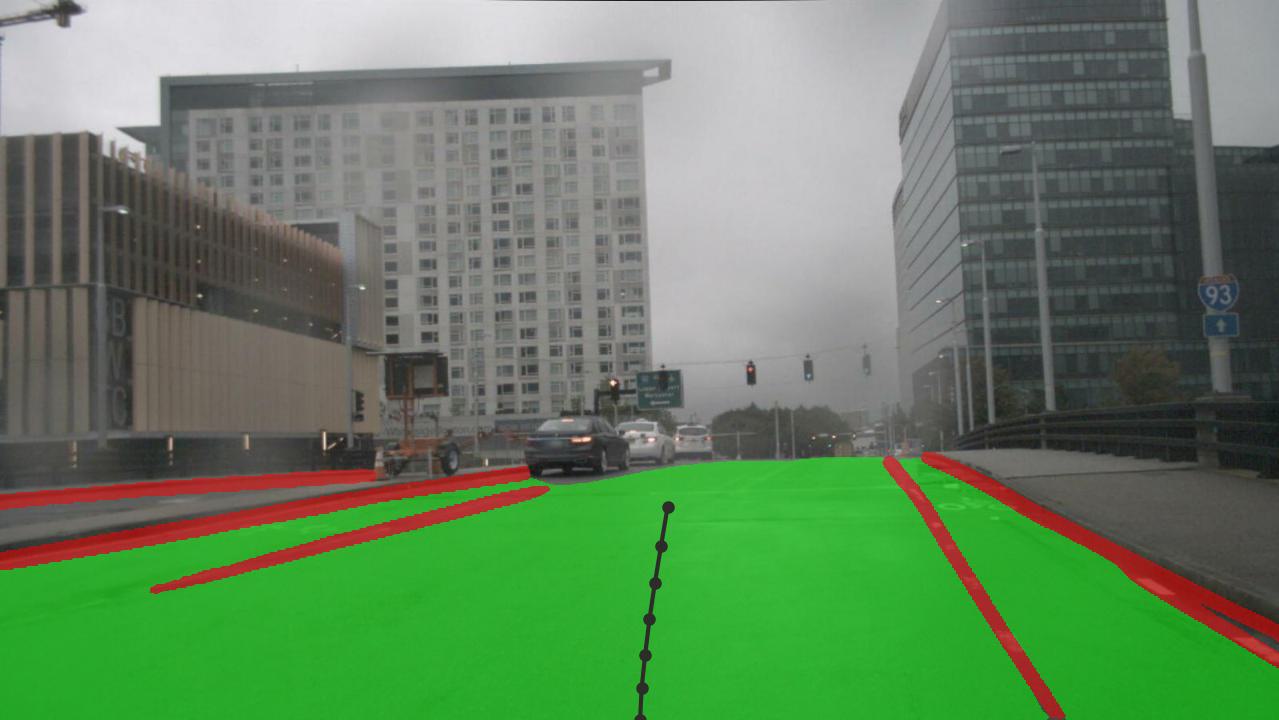}
    \end{subfigure}
    
    \caption{Visualization of the refined trajectory predictions across simulated, real-world's dark, and rainy scenes.}
    \label{fig:vis}
\end{figure*}

Fig. \ref{fig:vis} shows VDRive's refined trajectory output in nuScenes and the CARLA simulator. As VDRive learns to infer the drivable area directly from sensory input, the predicted trajectories conform to context- and geometry-aware driving corridors.

\subsection{Ablation Study}

In the ablation study, we report the closed-loop results of different dataset constructions and different refinement module designs in the Bench2Drive-mini set. Table \ref{tab:data} shows that our proposed dataset construction that combines real-world, synthetic, and simulated scenes provides the best performance in closed-loop tasks. In Table \ref{tab:refine}, we report the results of different dynamic fusion approaches, where cross attention achieved the best performance compared to the refinement based on LSTM and GRU \citep{graves2012lstm, dey2017gru}.

\begin{table}[!tb]
\centering
\caption{Ablation Study in Bench2Drive-mini set.}\vspace{-3mm}\label{tab:ablation}
\subcaptionbox{\textbf{Dataset Construction\vspace{-1mm}}\label{tab:data}}{
\begin{adjustbox}{width=0.49\linewidth}
\begin{tabular}{l|cc}
\toprule
Method                         & \textbf{Driving Score} $\uparrow$  & \textbf{Success Rate} $\uparrow$   \\ \midrule
nuScenes + Vista-generated          & 56.36     & 38.61  \\ 
Bench2Drive                         & 61.55       & 39.23 \\
nuScenes + Bench2Drive (\textbf{ours})          & \textbf{66.87}  & \textbf{51.66}  \\ \bottomrule
\end{tabular}
\end{adjustbox}
}
\subcaptionbox{\textbf{Refinement Module\vspace{-1mm}}\label{tab:refine}}{
\begin{adjustbox}{width=0.45\linewidth}
\begin{tabular}{l|cc}
\toprule
Method                         & \textbf{Driving Score} $\uparrow$  & \textbf{Success Rate} $\uparrow$     \\ \midrule
LSTM-based                         & 61.49       & 36.09 \\ 
GRU-based                         & 56.21       & 39.38 \\
Transformer-based (\textbf{ours})  & \textbf{69.51}       & \textbf{52.01} \\ \bottomrule
\end{tabular}
\end{adjustbox}
}
\vspace{-4mm}
\end{table}

%% file: iclr2026_conference.bib
@article{xu2025knowledge,
  title={A Knowledge-Driven Diffusion Policy for End-to-End Autonomous Driving Based on Expert Routing},
  author={Xu, Chengkai and Liu, Jiaqi and Guo, Yicheng and Hang, Peng and Sun, Jian},
  journal={arXiv preprint arXiv:2509.04853},
  year={2025}
}

@article{zhu2025internvl3,
      title={InternVL3: Exploring Advanced Training and Test-Time Recipes for Open-Source Multimodal Models}, 
      author={Jinguo Zhu and Weiyun Wang and Zhe Chen and Zhaoyang Liu and Shenglong Ye and Lixin Gu and Hao Tian and Yuchen Duan and Weijie Su and Jie Shao and Zhangwei Gao and Erfei Cui and Xuehui Wang and Yue Cao and Yangzhou Liu and Xingguang Wei and Hongjie Zhang and Haomin Wang and Weiye Xu and Hao Li and Jiahao Wang and Nianchen Deng and Songze Li and Yinan He and Tan Jiang and Jiapeng Luo and Yi Wang and Conghui He and Botian Shi and Xingcheng Zhang and Wenqi Shao and Junjun He and Yingtong Xiong and Wenwen Qu and Peng Sun and Penglong Jiao and Han Lv and Lijun Wu and Kaipeng Zhang and Huipeng Deng and Jiaye Ge and Kai Chen and Limin Wang and Min Dou and Lewei Lu and Xizhou Zhu and Tong Lu and Dahua Lin and Yu Qiao and Jifeng Dai and Wenhai Wang},
      year={2025},
      journal={arXiv preprint arXiv:2504.10479},
}

@article{jiang2025flowdrive,
  title={FlowDrive: Energy Flow Field for End-to-End Autonomous Driving},
  author={Jiang, Hao and Zhang, Zhipeng and Gao, Yu and Sun, Zhigang and Wang, Yiru and Heng, Yuwen and Wang, Shuo and Chai, Jinhao and Chen, Zhuo and Zhao, Hao and others},
  journal={arXiv preprint arXiv:2509.14303},
  year={2025}
}

@article{jiang2025irlvla,
  title={IRL-VLA: Training an Vision-Language-Action Policy via Reward World Model}, 
  author={Anqing Jiang and Yu Gao and Yiru Wang and Zhigang Sun and Shuo Wang and Yuwen Heng and Hao Sun and Shichen Tang and Lijuan Zhu and Jinhao Chai and Jijun Wang and Zichong Gu and Hao Jiang and Li Sun},
  year={2025},
  journal={arXiv preprint arXiv:2508.06571},
}

@ARTICLE{guidance,
  author={Li, Bowen and Wu, Tao and Yu, Youjin and Li, Junxiang},
  journal={IEEE Transactions on Vehicular Technology}, 
  title={End-to-End Autonomous Guidance Method Integrated With Mixture-of-Experts for Intelligent Vehicles}, 
  year={2025},
  volume={},
  number={},
  pages={1-15},
  doi={10.1109/TVT.2025.3603267}}

@article{liu2025reinforced,
  title={Reinforced Refinement with Self-Aware Expansion for End-to-End Autonomous Driving},
  author={Liu, Haochen and Li, Tianyu and Yang, Haohan and Chen, Li and Wang, Caojun and Guo, Ke and Tian, Haochen and Li, Hongchen and Li, Hongyang and Lv, Chen},
  journal={arXiv preprint arXiv:2506.09800},
  year={2025}
}

@article{gao2025rad,
  title={Rad: Training an end-to-end driving policy via large-scale 3dgs-based reinforcement learning},
  author={Gao, Hao and Chen, Shaoyu and Jiang, Bo and Liao, Bencheng and Shi, Yiang and Guo, Xiaoyang and Pu, Yuechuan and Yin, Haoran and Li, Xiangyu and Zhang, Xinbang and others},
  journal={arXiv preprint arXiv:2502.13144},
  year={2025}
}

@article{guo2025vdt,
  title={Vdt-auto: End-to-end autonomous driving with vlm-guided diffusion transformers},
  author={Guo, Ziang and Gubernatorov, Konstantin and Asfaw, Selamawit and Yagudin, Zakhar and Tsetserukou, Dzmitry},
  journal={arXiv preprint arXiv:2502.20108},
  year={2025}
}

@article{xu2024vlmad,
  title={Vlm-ad: End-to-end autonomous driving through vision-language model supervision},
  author={Xu, Yi and Hu, Yuxin and Zhang, Zaiwei and Meyer, Gregory P and Mustikovela, Siva Karthik and Srinivasa, Siddhartha and Wolff, Eric M and Huang, Xin},
  journal={arXiv preprint arXiv:2412.14446},
  year={2024}
}

@article{jiang2025VLAsurvey,
  title={A Survey on Vision-Language-Action Models for Autonomous Driving},
  author={Jiang, Sicong and Huang, Zilin and Qian, Kangan and Luo, Ziang and Zhu, Tianze and Zhong, Yang and Tang, Yihong and Kong, Menglin and Wang, Yunlong and Jiao, Siwen and others},
  journal={arXiv preprint arXiv:2506.24044},
  year={2025}
}

@article{hwang2024emma,
  title={Emma: End-to-end multimodal model for autonomous driving},
  author={Hwang, Jyh-Jing and Xu, Runsheng and Lin, Hubert and Hung, Wei-Chih and Ji, Jingwei and Choi, Kristy and Huang, Di and He, Tong and Covington, Paul and Sapp, Benjamin and others},
  journal={arXiv preprint arXiv:2410.23262},
  year={2024}
}

@article{li2024multimodal,
  title={Multimodal alignment and fusion: A survey},
  author={Li, Songtao and Tang, Hao},
  journal={arXiv preprint arXiv:2411.17040},
  year={2024}
}

@article{din2025VLAmani,
  title={Vision language action models in robotic manipulation: A systematic review},
  author={Din, Muhayy Ud and Akram, Waseem and Saoud, Lyes Saad and Rosell, Jan and Hussain, Irfan},
  journal={arXiv preprint arXiv:2507.10672},
  year={2025}
}

@article{xie2025vlmready,
  title={Are vlms ready for autonomous driving? an empirical study from the reliability, data, and metric perspectives},
  author={Xie, Shaoyuan and Kong, Lingdong and Dong, Yuhao and Sima, Chonghao and Zhang, Wenwei and Chen, Qi Alfred and Liu, Ziwei and Pan, Liang},
  journal={arXiv preprint arXiv:2501.04003},
  year={2025}
}

@article{zhou2025opendrivevla,
  title={Opendrivevla: Towards end-to-end autonomous driving with large vision language action model},
  author={Zhou, Xingcheng and Han, Xuyuan and Yang, Feng and Ma, Yunpu and Knoll, Alois C},
  journal={arXiv preprint arXiv:2503.23463},
  year={2025}
}

@article{zhou2025autovla,
  title={AutoVLA: A Vision-Language-Action Model for End-to-End Autonomous Driving with Adaptive Reasoning and Reinforcement Fine-Tuning},
  author={Zhou, Zewei and Cai, Tianhui and Zhao, Seth Z and Zhang, Yun and Huang, Zhiyu and Zhou, Bolei and Ma, Jiaqi},
  journal={arXiv preprint arXiv:2506.13757},
  year={2025}
}

@article{jiang2025diffvla,
  title={Diffvla: Vision-language guided diffusion planning for autonomous driving},
  author={Jiang, Anqing and Gao, Yu and Sun, Zhigang and Wang, Yiru and Wang, Jijun and Chai, Jinghao and Cao, Qian and Heng, Yuweng and Jiang, Hao and Dong, Yunda and others},
  journal={arXiv preprint arXiv:2505.19381},
  year={2025}
}

@article{cao2025fastdrivevla,
  title={FastDriveVLA: Efficient End-to-End Driving via Plug-and-Play Reconstruction-based Token Pruning},
  author={Cao, Jiajun and Zhang, Qizhe and Jia, Peidong and Zhao, Xuhui and Lan, Bo and Zhang, Xiaoan and Wei, Xiaobao and Chen, Sixiang and Li, Zhuo and Wang, Yang and others},
  journal={arXiv preprint arXiv:2507.23318},
  year={2025}
}

@article{chi2023diffusionpolicy,
  title={Diffusion policy: Visuomotor policy learning via action diffusion},
  author={Chi, Cheng and Xu, Zhenjia and Feng, Siyuan and Cousineau, Eric and Du, Yilun and Burchfiel, Benjamin and Tedrake, Russ and Song, Shuran},
  journal={The International Journal of Robotics Research},
  pages={02783649241273668},
  year={2023},
  publisher={SAGE Publications Sage UK: London, England}
}

@inproceedings{yang2024diff-es,
  title={Diffusion-ES: Gradient-free planning with diffusion for autonomous and instruction-guided driving},
  author={Yang, Brian and Su, Huangyuan and Gkanatsios, Nikolaos and Ke, Tsung-Wei and Jain, Ayush and Schneider, Jeff and Fragkiadaki, Katerina},
  booktitle={Proceedings of the IEEE/CVF conference on computer vision and pattern recognition},
  pages={15342--15353},
  year={2024}
}

@inproceedings{peebles2023DiT,
  title={Scalable diffusion models with transformers},
  author={Peebles, William and Xie, Saining},
  booktitle={Proceedings of the IEEE/CVF International Conference on Computer Vision},
  pages={4195--4205},
  year={2023}
}

@article{liu2024rdt-1b,
  title={Rdt-1b: a diffusion foundation model for bimanual manipulation},
  author={Liu, Songming and Wu, Lingxuan and Li, Bangguo and Tan, Hengkai and Chen, Huayu and Wang, Zhengyi and Xu, Ke and Su, Hang and Zhu, Jun},
  journal={arXiv preprint arXiv:2410.07864},
  year={2024}
}

@inproceedings{rombach2022latentDiff,
  title={High-resolution image synthesis with latent diffusion models},
  author={Rombach, Robin and Blattmann, Andreas and Lorenz, Dominik and Esser, Patrick and Ommer, Bj{\"o}rn},
  booktitle={Proceedings of the IEEE/CVF conference on computer vision and pattern recognition},
  pages={10684--10695},
  year={2022}
}

@inproceedings{yang2023diffrep,
  title={Diffusion model as representation learner},
  author={Yang, Xingyi and Wang, Xinchao},
  booktitle={Proceedings of the IEEE/CVF International Conference on Computer Vision},
  pages={18938--18949},
  year={2023}
}

@article{song2025breaking,
  title={Breaking Imitation Bottlenecks: Reinforced Diffusion Powers Diverse Trajectory Generation},
  author={Song, Ziying and Liu, Lin and Pan, Hongyu and Liao, Bencheng and Guo, Mingzhe and Yang, Lei and Zhang, Yongchang and Xu, Shaoqing and Jia, Caiyan and Luo, Yadan},
  journal={arXiv preprint arXiv:2507.04049},
  year={2025}
}

@inproceedings{caesar2020nuscenes,
  title={nuscenes: A multimodal dataset for autonomous driving},
  author={Caesar, Holger and Bankiti, Varun and Lang, Alex H and Vora, Sourabh and Liong, Venice Erin and Xu, Qiang and Krishnan, Anush and Pan, Yu and Baldan, Giancarlo and Beijbom, Oscar},
  booktitle={Proceedings of the IEEE/CVF conference on computer vision and pattern recognition},
  pages={11621--11631},
  year={2020}
}

@inproceedings{gao2024vista,
 title={Vista: A Generalizable Driving World Model with High Fidelity and Versatile Controllability}, 
 author={Shenyuan Gao and Jiazhi Yang and Li Chen and Kashyap Chitta and Yihang Qiu and Andreas Geiger and Jun Zhang and Hongyang Li},
 booktitle={Advances in Neural Information Processing Systems (NeurIPS)},
 year={2024}
}

@article{zeng2025futuresightdrive,
  title={FutureSightDrive: Thinking Visually with Spatio-Temporal CoT for Autonomous Driving},
  author={Zeng, Shuang and Chang, Xinyuan and Xie, Mengwei and Liu, Xinran and Bai, Yifan and Pan, Zheng and Xu, Mu and Wei, Xing},
  journal={arXiv preprint arXiv:2505.17685},
  year={2025}
}

@article{jia2024bench2drive,
  title={Bench2drive: Towards multi-ability benchmarking of closed-loop end-to-end autonomous driving},
  author={Jia, Xiaosong and Yang, Zhenjie and Li, Qifeng and Zhang, Zhiyuan and Yan, Junchi},
  journal={Advances in Neural Information Processing Systems},
  volume={37},
  pages={819--844},
  year={2024}
}

@article{mentzer2023finite,
  title={Finite scalar quantization: Vq-vae made simple},
  author={Mentzer, Fabian and Minnen, David and Agustsson, Eirikur and Tschannen, Michael},
  journal={arXiv preprint arXiv:2309.15505},
  year={2023}
}

@article{bai2025qwen2,
  title={Qwen2. 5-vl technical report},
  author={Bai, Shuai and Chen, Keqin and Liu, Xuejing and Wang, Jialin and Ge, Wenbin and Song, Sibo and Dang, Kai and Wang, Peng and Wang, Shijie and Tang, Jun and others},
  journal={arXiv preprint arXiv:2502.13923},
  year={2025}
}

@article{chen2024diffusionoffline,
  title={Diffusion policies creating a trust region for offline reinforcement learning},
  author={Chen, Tianyu and Wang, Zhendong and Zhou, Mingyuan},
  journal={Advances in Neural Information Processing Systems},
  volume={37},
  pages={50098--50125},
  year={2024}
}

@article{graves2012lstm,
  title={Long short-term memory},
  author={Graves, Alex},
  journal={Supervised sequence labelling with recurrent neural networks},
  pages={37--45},
  year={2012},
  publisher={Springer}
}

@inproceedings{dey2017gru,
  title={Gate-variants of gated recurrent unit (GRU) neural networks},
  author={Dey, Rahul and Salem, Fathi M},
  booktitle={2017 IEEE 60th international midwest symposium on circuits and systems (MWSCAS)},
  pages={1597--1600},
  year={2017},
  organization={IEEE}
}

@inproceedings{hu2021safeff,
  title={Safe local motion planning with self-supervised freespace forecasting},
  author={Hu, Peiyun and Huang, Aaron and Dolan, John and Held, David and Ramanan, Deva},
  booktitle={Proceedings of the IEEE/CVF Conference on Computer Vision and Pattern Recognition},
  pages={12732--12741},
  year={2021}
}

@inproceedings{kh2022differentiable,
  title={Differentiable raycasting for self-supervised occupancy forecasting},
  author={Khurana, Tarasha and Hu, Peiyun and Dave, Achal and Ziglar, Jason and Held, David and Ramanan, Deva},
  booktitle={European Conference on Computer Vision},
  pages={353--369},
  year={2022},
  organization={Springer}
}

@inproceedings{hu2022st-p3,
  title={St-p3: End-to-end vision-based autonomous driving via spatial-temporal feature learning},
  author={Hu, Shengchao and Chen, Li and Wu, Penghao and Li, Hongyang and Yan, Junchi and Tao, Dacheng},
  booktitle={European Conference on Computer Vision},
  pages={533--549},
  year={2022},
  organization={Springer}
}

@inproceedings{hu2023uniAD,
  title={Planning-oriented autonomous driving},
  author={Hu, Yihan and Yang, Jiazhi and Chen, Li and Li, Keyu and Sima, Chonghao and Zhu, Xizhou and Chai, Siqi and Du, Senyao and Lin, Tianwei and Wang, Wenhai and others},
  booktitle={Proceedings of the IEEE/CVF Conference on Computer Vision and Pattern Recognition},
  pages={17853--17862},
  year={2023}
}

@article{mao2023gpt,
  title={Gpt-driver: Learning to drive with gpt},
  author={Mao, Jiageng and Qian, Yuxi and Ye, Junjie and Zhao, Hang and Wang, Yue},
  journal={arXiv preprint arXiv:2310.01415},
  year={2023}
}

@inproceedings{pan2024vlp,
  title={VLP: Vision Language Planning for Autonomous Driving},
  author={Pan, Chenbin and Yaman, Burhaneddin and Nesti, Tommaso and Mallik, Abhirup and Allievi, Alessandro G and Velipasalar, Senem and Ren, Liu},
  booktitle={Proceedings of the IEEE/CVF Conference on Computer Vision and Pattern Recognition},
  pages={14760--14769},
  year={2024}
}

@inproceedings{huang2024rda,
  title={Making large language models better planners with reasoning-decision alignment},
  author={Huang, Zhijian and Tang, Tao and Chen, Shaoxiang and Lin, Sihao and Jie, Zequn and Ma, Lin and Wang, Guangrun and Liang, Xiaodan},
  booktitle={European Conference on Computer Vision},
  pages={73--90},
  year={2024},
  organization={Springer}
}

@article{tian2024drivevlm,
  title={Drivevlm: The convergence of autonomous driving and large vision-language models},
  author={Tian, Xiaoyu and Gu, Junru and Li, Bailin and Liu, Yicheng and Wang, Yang and Zhao, Zhiyong and Zhan, Kun and Jia, Peng and Lang, Xianpeng and Zhao, Hang},
  journal={arXiv preprint arXiv:2402.12289},
  year={2024}
}

@article{wang2024hedrive,
  title={HE-Drive: Human-Like End-to-End Driving with Vision Language Models},
  author={Wang, Junming and Zhang, Xingyu and Xing, Zebin and Gu, Songen and Guo, Xiaoyang and Hu, Yang and Song, Ziying and Zhang, Qian and Long, Xiaoxiao and Yin, Wei},
  journal={arXiv preprint arXiv:2410.05051},
  year={2024}
}

@article{jia2025drivetransformer,
      title={DriveTransformer: Unified Transformer for Scalable End-to-End Autonomous Driving}, 
      author={Xiaosong Jia and Junqi You and Zhiyuan Zhang and Junchi Yan},
      journal={arXiv preprint arXiv:2503.07656},
      year={2025}
}

@article{zhai2023ADMLP,
  title={Rethinking the Open-Loop Evaluation of End-to-End Autonomous Driving in nuScenes},
  author={Zhai, Jiang-Tian and Feng, Ze and Du, Jihao and Mao, Yongqiang and Liu, Jiang-Jiang and Tan, Zichang and Zhang, Yifu and Ye, Xiaoqing and Wang, Jingdong},
  journal={arXiv preprint arXiv:2305.10430},
  year={2023}
}

@inproceedings{hu2023planning,
  title={Planning-oriented autonomous driving},
  author={Hu, Yihan and Yang, Jiazhi and Chen, Li and Li, Keyu and Sima, Chonghao and Zhu, Xizhou and Chai, Siqi and Du, Senyao and Lin, Tianwei and Wang, Wenhai and others},
  booktitle={CVPR},
  pages={17853--17862},
  year={2023}
}

@article{jiang2023vad,
  title={VAD: Vectorized Scene Representation for Efficient Autonomous Driving},
  author={Jiang, Bo and Chen, Shaoyu and Xu, Qing and Liao, Bencheng and Chen, Jiajie and Zhou, Helong and Zhang, Qian and Liu, Wenyu and Huang, Chang and Wang, Xinggang},
  journal={ICCV},
  year={2023}
}

@inproceedings{wu2022trajectoryguided,
 title={Trajectory-guided Control Prediction for End-to-end Autonomous Driving: A Simple yet Strong Baseline}, 
 author={Penghao Wu and Xiaosong Jia and Li Chen and Junchi Yan and Hongyang Li and Yu Qiao},
 booktitle={NeurIPS},
 year={2022},
}

@inproceedings{jia2023thinktwice,
  title={Think Twice before Driving: Towards Scalable Decoders for End-to-End Autonomous Driving},
  author={Jia, Xiaosong and Wu, Penghao and Chen, Li and Xie, Jiangwei and He, Conghui and Yan, Junchi and Li, Hongyang},
  booktitle={CVPR},
  year={2023}
}

@inproceedings{jia2023driveadapter,
  title={DriveAdapter: Breaking the Coupling Barrier of Perception and Planning in End-to-End Autonomous Driving},
  author={Jia, Xiaosong and Gao, Yulu and Chen, Li and Yan, Junchi and Liu, Patrick Langechuan and Li, Hongyang},
  booktitle={ICCV},
  year={2023}
}

@article{wang2025cogad,
      title={CogAD: Cognitive-Hierarchy Guided End-to-End Autonomous Driving}, 
      author={Zhennan Wang and Jianing Teng and Canqun Xiang and Kangliang Chen and Xing Pan and Lu Deng and Weihao Gu},
      year={2025},
      journal={arXiv preprint arXiv:2505.21581},
}

@article{lu2025real-ad,
      title={ReAL-AD: Towards Human-Like Reasoning in End-to-End Autonomous Driving}, 
      author={Yuhang Lu and Jiadong Tu and Yuexin Ma and Xinge Zhu},
      year={2025},
      journal={arXiv preprint arXiv:2507.12499},
}
